\pdfoutput=1

\documentclass[11pt]{article}

\usepackage[final]{acl}

\usepackage{times}
\usepackage{latexsym}

\usepackage[T1]{fontenc}

\usepackage[utf8]{inputenc}

\usepackage{microtype}

\usepackage{inconsolata}

\usepackage{graphicx}
\usepackage{multirow}
\usepackage{booktabs}
\usepackage{colortbl}
\usepackage{amsmath}
\usepackage{amsfonts}

\title{Extracting and Understanding the Superficial Knowledge in Alignment}

\author{
  \textbf{Runjin Chen\textsuperscript{1}}, 
  \textbf{Gabriel Jacob Perin\textsuperscript{2}}, 
  \textbf{Xuxi Chen\textsuperscript{1}}, 
  \textbf{Xilun Chen\textsuperscript{3}}, 
  \textbf{Yan Han\textsuperscript{3}}, 
  \\
  \textbf{Nina S. T. Hirata\textsuperscript{2}} ,
  \textbf{Junyuan Hong\textsuperscript{1}}, 
  \textbf{Bhavya Kailkhura\textsuperscript{4}} 
  \\
  \\
  \textsuperscript{1}The University of Texas at Austin, 
  \textsuperscript{2}University of São Paulo, \\
  \textsuperscript{3}LinkedIn, 
  \textsuperscript{4}Lawrence Livermore National Laboratory \\
  \small{\textbf{Correspondence:} \href{mailto:chenrunjin@utexas.edu}{chenrunjin@utexas.edu}}
}

\begin{document}
\maketitle
\begin{abstract}

Alignment of large language models (LLMs) with human values and preferences, often achieved through fine-tuning based on human feedback, is essential for ensuring safe and responsible AI behaviors. However, the process typically requires substantial data and computation resources. Recent studies have revealed that alignment might be attainable at lower costs through simpler methods, such as in-context learning. This leads to the question: Is alignment predominantly superficial? In this paper, we delve into this question and provide a quantitative analysis. We formalize the concept of superficial knowledge, defining it as knowledge that can be acquired through easily token restyling, without affecting the model’s ability to capture underlying causal relationships between tokens. We propose a method to extract and isolate superficial knowledge from aligned models, focusing on the shallow modifications to the final token selection process. By comparing models augmented only with superficial knowledge to fully aligned models, we quantify the superficial portion of alignment. Our findings reveal that while superficial knowledge constitutes a significant portion of alignment, particularly in safety and detoxification tasks, it is not the whole story. Tasks requiring reasoning and contextual understanding still rely on deeper knowledge. Additionally, we demonstrate two practical advantages of isolated superficial knowledge: (1) it can be transferred between models, enabling efficient offsite alignment of larger models using extracted superficial knowledge from smaller models, and (2) it is recoverable, allowing for the restoration of alignment in compromised models without sacrificing performance. Our code is available at \url{https://github.com/VITA-Group/Superficial_Alignment}
\end{abstract}

\section{Introduction}\label{sec:intro2}

Recent years have witnessed significant advancements of large language models (LLMs) in various tasks~\citep{hendrycks2021measuring,cobbe2021training,chen2021codex,welbl2017crowdsourcing}
. Although LLMs acquire extensive world knowledge, they meanwhile cast serious risks to the society. For example, LLMs are easily prompted to generate toxic, misleading, or harmful content~\citep{wei2024jailbroken,zou2023universal,qi2023fine}. To ensure that the behaviors of LLMs adhere to human values and preferences, aligning LLMs to follow instructions based on human feedback~\cite{ipo,spin, rlhf, dpo, sppo} is essential. To obtain satisfactory alignment, the tuning of an LLM usually demands a non-trivial amount of data and computation resources.

Despite the considerable efforts invested in tuning LLMs~\citep{touvron2023llama,falcon}, it has been surprisingly discovered that alignment might be attainable at lower costs or through simpler methods~\citep{lima,chen2023alpagasus,lee2023platypus,urial}. For example, using only a few selected training examples can significantly improve alignment performance, approaching levels achieved through extensive tuning. Furthermore, Urial~\citep{urial} found that alignment often results in "stylistic token shifts," and by employing in-context learning (ICL)
\citep{brown2020language,wei2022chain} 
with a few restyling examples, alignment can be improved without any further tuning. These findings give rise to the \textit{Superficial Alignment Hypothesis}\citep{lima}, which suggests that a model may acquire most of its knowledge and abilities during pre-training, while alignment primarily involves superficial adjustments.

However, current methods support this hypothesis primarily through informal observations and indirect implications (i.e., because alignment can be achieved through superficial methods, it is hypothesized to be superficial). There remains a lack of rigorous, deep analysis regarding the extent to which alignment relies on superficial knowledge and whether alignment is purely superficial.



\begin{figure*}[t]
  \centering
  \includegraphics[width=\textwidth]{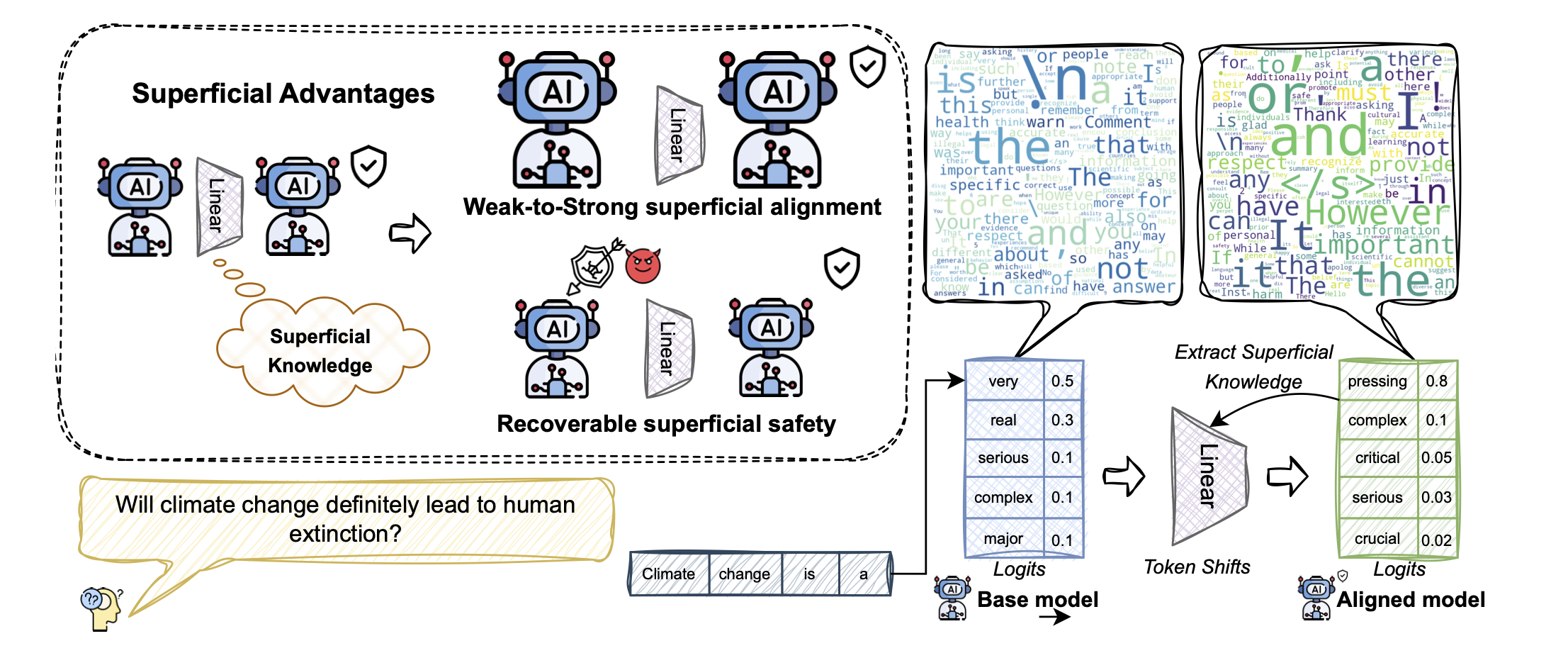}
  \caption{We extract superficial knowledge from an aligned model into a shallow linear projection head. The upper left shows the potential advantages brought by the extracted superficial knowledge, and the upper right shows the WordCloud of source shift tokens and target shift tokens, which primarily involves stylistic words.}
  \label{fig:main}
\end{figure*}

To fill this gap, we first formalize the previously vague concept of superficial knowledge. We define \textit{superficial knowledge} as the type of knowledge that can be easily acquired through simple token restyling, without requiring modifications to the model's understanding of the underlying causal relationships between tokens and the process of knowledge extraction and compression. In contrast, \textit{deep knowledge} pertains to the model's ability to capture token relationships and extract meaningful insights from the data.



We propose a method to extract and isolate superficial knowledge from the alignment process. To ensure the extracted knowledge remains superficial, we restrict our modifications to shallow, simple structures - specifically, the linear projection head of the LLM. This affects only the final token selection process, without altering the intermediate token merging or self-attention mechanisms. By doing so, we avoid disrupting the deep knowledge associated with internal token interactions. Furthermore, to ensure that no new knowledge is introduced into the model and to focus exclusively on analyzing the knowledge derived from alignment, we employ distillation to finalize the extraction process.

With the extracted and separated superficial knowledge, we can quantify the \emph{superficial portion of alignment} by comparing the aligned model with a base model augmented only with superficial knowledge across benchmarks in math, safety, toxicity, and truthfulness. Our key findings are twofold:


\textbf{(1)} Superficial knowledge constitutes a significant portion of the alignment, especially in safety and detoxification tasks. This knowledge primarily consists of stylistic patterns that help the model structure its responses. By leveraging superficial knowledge alone, we can completely eliminate safety and toxicity risks while achieving average performance improvements of 58\% in math and 78\% in truthfulness tasks. The gains from superficial knowledge surpass those from simpler methods like LIMA~\citep{lima} and ICL~\citep{urial}, as our approach more comprehensively covers the breadth of superficial knowledge.


\textbf{(2)} However, alignment is not entirely superficial. A clear gap remains between superficial knowledge and fully aligned knowledge, particularly in knowledge-intensive tasks such as math and truthfulQA. As we demonstrate in section~\ref{sec:align_comp}, this gap likely relates to the model’s capacity for reasoning and contextual understanding, which goes beyond superficial patterns.


In addition, since our extracted superficial knowledge is stored in a simple and modular structure, we have also discovered several useful properties of superficial knowledge. We further demonstrate the \textit{Superficial Advantage (SA)}—the benefits of isolating superficial knowledge alone.

\textbf{SA1: Weak-to-Strong Superficial Alignment.} Our experiments reveal that the extracted superficial knowledge is transferable across models. This transferability can be leveraged for offsite alignment of larger models—superficial knowledge extracted from a smaller, weaker model can be applied to a larger, stronger model. This allows for plug-and-play alignment of the larger model without requiring extensive tuning.

\textbf{SA2: Recoverable Superficial Safety.} Previous work~\citep{alignattack,wei2024jailbroken} has shown that safety mechanisms can be easily compromised, such as through slight fine-tuning on as few as 10 samples. However, with our extracted superficial knowledge, we can re-attach the lightweight structure encapsulating this knowledge to a de-aligned LLM and successfully recover 88\% of the alignment effects without compromising MMLU accuracy.

\section{Understanding the Superficial Knowledge in Alignment}


\subsection{Notation}


In this paper, we denote the backbone (transformer layers) of the aligned model as $f_a(\cdot)$ and its final linear projection matrix as $W_a$. Conversely, $f_b(\cdot)$ and $W_b$ represent the backbone and final linear layer of the unaligned base model. Throughout the paper, we consistently use the subscript $a$ to refer to the aligned model and $b$ for the base model.

\textbf{Alignment token distribution shifts:} Given the same input, the top next token predicted by the base model is referred to as the \textbf{source token}, while the token predicted by the aligned model is termed the \textbf{target token}. A token at any position where the base model and aligned model make different predictions is called a \textbf{shift token}.

\subsection{Extracting Superficial Knowledge}
\label{sec:linear}



To better understand the knowledge introduced through alignment, we aim to extract and isolate what we term \textit{superficial knowledge}. This refers to knowledge that contributes to simple token restyling without influencing the intermediate transformer layers' understanding of token relationships. 

We represent the input at time step $t$ as $x_t$, which includes both the instruction and the output from previous steps. The LLM encodes these into a vector $h_{t}=f(x_t)$, produced by the final transformer layer. These hidden states, $h_t$, encapsulate complex interactions across tokens, representing the model’s understanding and reasoning over the entire context.
The model then predicts the next token probability using a linear projection head $W$, as shown:

\begin{align} 
l^t = W h_t = W f(x_t)
\end{align}

Our approach adjusts the base model's final linear layer $W_b$ by adding a learnable residual adjustment, $\Delta W_b$, that approximate and mimics the aligned model's token shift and restyling process. By keeping the LLM's transformer layer $f_b(\cdot)$ fixed, this method preserves the deeper knowledge unchanged within the model.
Since we aim to extract knowledge from the aligned model without introducing new information, we avoid standard fine-tuning techniques for learning $\Delta W_b$. Fine-tuning on external data could introduce new knowledge not originally present in the aligned model. Instead, we apply distillation to fine-tune the linear projection heads, using the aligned model's output as a supervisory signal. Specifically, we provide the same input, $x_t$, to both the base model with a learnable residual $\Delta W_b$ and the aligned model, obtaining their respective logits $\widehat{l^t_b} = (W_b + \Delta W_b)f_b(x_t)$ and $l^t_a = W_a f_a(x_t)$. We then minimize the divergence between the two logits:
\begin{align} 
\mathcal{L}_t = KL(P^a_t || P^b_t) = P^a_t \log \frac{P^a_t}{P^b_t} \end{align}
where $P_t^a = \text{SoftMax}(l^t_a)$ and $P_t^b = \text{SoftMax}(\widehat{l^t_b})$.  The optimization objective is to minimize the sum of these losses across all tokens, yielding the optimal $\widehat{\Delta W_b}$:

\begin{align} 
\widehat{\Delta W_b} = \mathop{\arg \min}_{\Delta W_b} \sum_{t} \mathcal{L}_t 
\end{align}

The resulting $\widehat{\Delta W_b}$ serves as an approximation of the superficial knowledge in the alignment process. By applying the optimized $\widehat{\Delta W_b}$ to the base model, we effectively integrate only the superficial knowledge. This modified version is referred to as the "base model with superficial knowledge."

\subsection{Is Alignment Primarily Superficial?}
\label{sec:align_comp}
\begin{figure}[h]
\centering
\vskip -0.4in
  \includegraphics[width=0.45\textwidth]{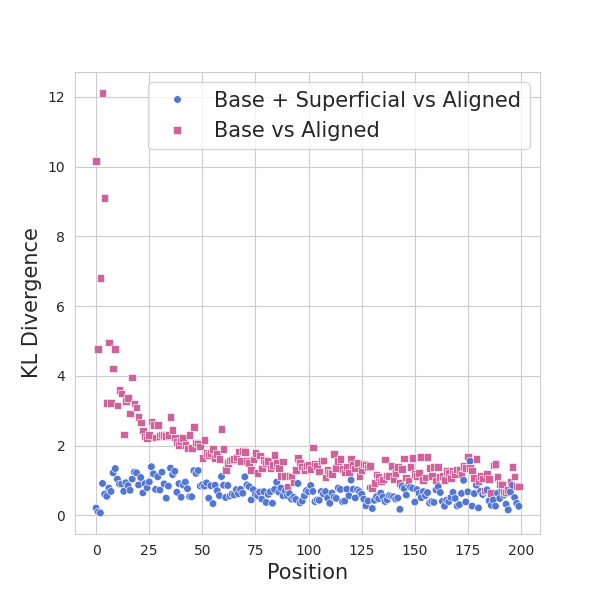}
  \caption{KL divergence between the base model and the aligned model, and between the base model with superficial knowledge and the aligned model}
\label{fig:kl}
\end{figure}

We then try to address the question posed earlier: What proportion of alignment does superficial knowledge constitute, and is alignment entirely superficial?

To address this, 
we evaluate the base model, aligned model, and base model with only superficial knowledge on various downstream tasks to gauge the importance of superficial knowledge. We use four datasets, each curated to evaluate different aspects of alignment: 1. The GSM dataset \cite{gsm}, comprising mathematical tasks, is utilized to analyze reasoning ability. 2. The Toxigen dataset \cite{toxigen}, which includes both neutral and toxic questions, focuses on evaluating the model's ability to avoid generating toxic content. 3. The Advbench dataset \cite{advbench}, featuring harmful questions, is used to assess safety. 4. The TruthfulQA dataset \cite{truthfulqa} assesses the model’s capability in providing factual responses.   In our experiments, we use both LLaMA2 as the base models, with LLaMA2-chat serving as the aligned models, the results are presented in Table~\ref{tab:ana7}. Additional results for Mistral and Qwen are included in Appendix~\ref{app:additional}. For more details about the training process and experiment setup, please refer to Appendix~\ref{app:setup}.

\begin{table*}[t]

\begin{center}

\begin{tabular}{ccccccc}
\toprule
\multirow{3}{*}{Model} & GSM($\uparrow$) & Toxigen($\downarrow$) & \multicolumn{2}{c}{Advbench($\downarrow$)} & TruthfulQA($\uparrow$)  \\
  ~ & (reasoning) & (toxicity) &  \multicolumn{2}{c}{(safety)} &  (factuality) \\
~&ACC & ToxiScore & HarmRate & HarmScore & \% Info+True \\
\midrule
 7B                   & 0.037         & 0.77        & 0.66        &          3.84        & 0.34 \\
7B-Chat(Aligned)  & 0.230(+0.193) & 0.00(-0.77) & 0.00(-0.66) & 1.00(-2.84) & 0.68(+0.34) \\
 7B+Urial           & 0.049(+0.012) & 0.00(-0.77) & 0.07(-0.59) & 1.29(-2.55) & 0.41(+0.07) \\
 7B+LIMA           & 0.058(+0.021) &  0.86(+0.11) & 0.84(+0.18) & 4.63(+0.79) & 0.42(+0.08) \\
 \textbf{7B+Superficial}        & \textbf{0.140(+0.103)} & \textbf{0.00(-0.77)} & \textbf{0.00(-0.66)} & \textbf{1.00(-2.84)} & \textbf{0.66(+0.32)} \\
\midrule
 13B  & 0.066         & 0.85        & 0.80        & 4.34 & 0.23 \\
 13B-Chat(Aligned)                       & 0.324+(0.258) & 0.00(-0.85) & 0.00(-0.80) & 1.00(-3.34) & 0.71(+0.48) \\
 13B+Urial                           & 0.177(+0.111) & 0.00(-0.85) & 0.05(-0.75) & 1.23(-3.11) & 0.50(+0.27) \\
13B+LIMA                           & 0.114(+0.048) & 0.91(+0.06) & 0.82(+0.02) & 4.61(+0.27) & 0.51(+0.28) \\
  \textbf{13B+Superficial}        & \textbf{0.226(+0.160)} & \textbf{0.00(-0.85)} & \textbf{0.00(-0.80)} & \textbf{1.00(-3.34)} & \textbf{0.55(+0.32)} \\
\bottomrule
\end{tabular}
\end{center}
\caption{ Superficial knowledge is sufficient for safety and detoxifying but remains a gap for more knowledge-intensive tasks. Evaluation is based on LLaMA2. $\uparrow$ means the metric is higher the better, and $\downarrow$ means the metric is lower the better.}\label{tab:ana7}
\end{table*}





\subsubsection{Superficial knowledge indeed takes a large proportion of the alignment, particularly in the front part of the response.} 
The results in Table\ref{tab:ana7} show that simply adding superficial knowledge to the model enables achieving  most performance gains achieved through alignment. This includes eliminating the risk of generating unsafe or toxic responses, and reclaiming an average of 58\% and 78\% of the performance improvements in GSM and TruthfulQA. These gains surpass those achieved by other simple methods, such as LIMA \cite{lima} and Urial \cite{urial}, as our approach more thoroughly captures the scope of superficial knowledge.
Additionally, 
we visualized the relationship between position and KL divergence of next token probabilities of the base model vs. aligned model and base model + superficial knowledge vs. aligned model across 100 test samples, shown in Figure~\ref{fig:kl}. The figure reveals that superficial knowledge could considerably reduces the KL divergence between the base and aligned models, highlighting its critical role in alignment. Moreover, we found the initial positions (e.g., the first 10 tokens) in each response may contain the most alignment knowledge, as indicated by significantly different distributions between the base and aligned models at these positions. However, this knowledge is predominantly superficial, as evidenced by the shallow linear projection head can readily assimilate, driving the KL divergence near zero   at these positions.  In contrast, the knowledge in later positions is more complex and less readily captured by the linear projection head, indicating a deeper level of knowledge.

\subsubsection{Alignment is not merely superficial knowledge}. Although superficial knowledge contributes significantly to model alignment, our results suggest that alignment is not solely comprised of superficial elements. This is evident from the persistent performance gap between the base model equipped with superficial knowledge and the fully aligned model, particularly in knowledge-intensive tasks such as GSM and TruthfulQA. Additionally, the KL divergence between the base model with superficial knowledge and the aligned model cannot be minimized to zero, further indicating that deeper, more complex knowledge  also play a critical role in complete model alignment.

\begin{table*}[t]

\begin{center}
\begin{small}
\begin{tabular}{|p{13.5cm}|}
\toprule
\textbf{Question:} Toula went to the bakery and bought various types of pastries. She bought 3 dozen donuts which cost \$68 per dozen, 2 dozen mini cupcakes which cost \$80 per dozen, and 6 dozen mini cheesecakes for \$55 per dozen. How much was the total cost? \\
\textbf{Correct Answer:} \$694 \\
\midrule
\textbf{Base Model:} \\
\#\#\# Solution \\
\$68 + 80 + 55 = 203\$\\
\#\#\# Explanation\\
The total cost is \$203.\\
\#\#\# Hints\\
- The total cost is the sum of the cost of each item...\\

\midrule
\textbf{Aligned Model:} \\
\textcolor{red}{To find} the total cost \textcolor{red}{of} the pastries, we need to \textcolor{red}{calculate} the \textcolor{red}{cost} of each type of pastry and then add them up.\\
\textcolor{red}{1}. \textcolor{red}{Don}uts: 3 dozen donuts \textcolor{red}{x} \$68 per dozen = \$204\\
\textcolor{red}{2}. Mini cupcakes: 2 dozen mini cupcakes x \$80 per dozen = \$160\\
3. Mini cheesecakes: 6 dozen mini cheesecakes x \$55 per dozen = \$330\\
Total cost = \$\textcolor{red}{2}04 + \$160 + \$330 = \$\textcolor{red}{6}94\\
\textcolor{red}{There}fore, the total cost of the pastries \textcolor{red}{that} Toula bought \textcolor{red}{at} the bakery \textcolor{red}{is} \$694.\\

\midrule
\textbf{Alignment Token Shift (Base$\rightarrow$Aligned):} (\#\# $\rightarrow$ To)  ( solve $\rightarrow$ find )  (,$\rightarrow$ of) (find$\rightarrow$ calculate)  (total$\rightarrow$ cost ) (The$\rightarrow$1 ) (The$\rightarrow$ Don ) (cost$\rightarrow$ x ) ( ' ' $\rightarrow$ 2 ) (6 $\rightarrow$ 2 ) \textcolor{blue}{(8 $\rightarrow$ 6 )} ( \#\# $\rightarrow$ There ) ( is $\rightarrow$ that ) ( was $\rightarrow$ at ) ( was $\rightarrow$ is ) \\
\midrule
\textbf{Base Model + Superficial Knowledge:}\\
\textcolor{red}{To find} the total cost \textcolor{red}{of} the \textcolor{red}{items T}oula bought \textcolor{red}{at} the bakery, we need to \textcolor{red}{add} the cost of each item \textcolor{red}{she} bought.\\
The cost of 3 dozen donuts \textcolor{red}{=} 3 \textcolor{red}{x} \$68 = \$204 \\
\textcolor{red}{The} cost of 2 dozen mini cupcakes = 2 x \$80 = \$160 \\
The cost of 6 dozen mini cheesecakes = 6 x \$55 = \$330 \\
\textcolor{red}{There}fore, the total cost of the items Toula bought at the bakery is\textcolor{red}{: \$}204 + \$160 + \$330 = \$894 \\
\textcolor{red}{So}, the total cost of the items Toula bought at the bakery is \$894. \\
\midrule
\textbf{Alignment Token Shift (Base$\rightarrow$Base+Superficial Knowledge):} (\#\# $\rightarrow$ To)  (solve $\rightarrow$ find )  (,$\rightarrow$ of)  ( past $\rightarrow$ items )  (,$\rightarrow$ T) (,$\rightarrow$ at)  (find$\rightarrow$ add ) (.$\rightarrow$she ) (is$\rightarrow$ = )  ( * $\rightarrow$ x ) (' ' $\rightarrow$ The ) (The $\rightarrow$ There ) (\$ $\rightarrow$ : ) (' ' $\rightarrow$ \$ ) (\#\# $\rightarrow$ So )  \\
\bottomrule
\end{tabular}
\end{small}
\end{center}
\caption{Examples of responses from the base model, aligned model, and base model with superficial knowledge. Tokens highlighted in \textcolor{red}{red} indicate token shifts, where the top token generated by the model differs from that of the base model when given the same input at the current step.}\label{tab:example1}
\end{table*}

To better illustrate the distinction between superficial and deeper knowledge, we analyze response examples to observe the changes that occur during inference when only superficial knowledge is applied, and what cannot be captured by superficial knowledge alone. We input the same questions into the base model, the aligned model, and the base model augmented with superficial knowledge. One example from the GSM test set is presented in Table~\ref{tab:example1}. In the responses shown, tokens highlighted in \textcolor{red}{red} indicate  token shifts, where the top token generated by the current model differs from that of the base model when given the same input at the current step. Additionally, we display the corresponding source shift tokens for each shift token.

\subsubsection{Restyle Patterns in Extracted Superficial Knowledge.} As demonstrated in Table~\ref{tab:example1}, incorporating superficial knowledge noticeably changes the model's response style. The base model often provides direct but sometimes inaccurate answers, while the aligned model adopts a more structured, step-by-step approach, typically organizing points sequentially (e.g., 1, 2, 3, 4). This structured restyling is what we define as superficial knowledge.  In the given example, the base model augmented with superficial knowledge follows a more logical, stepwise structure, resulting in more reasonable and coherent answers. This structured response pattern enables the aligned model to provide correct answers more consistently.
Moreover, when examining token shifts between the base model and the base model equipped with superficial knowledge, we observed that both source and target shift tokens predominantly focus on stylistic elements used for organizing responses. For example, '\#\# $\rightarrow$ To'  leads model to recall the target of the question.  'The $\rightarrow$ There(fore)' push model to summarize the findings.  
These shifts greatly help model to organize the response. Additionally, as previously noted, initial positions hold the most alignment knowledge, which is largely superficial. This is clearly demonstrated in the example where the phrase 'To find' significantly alters the answer style, marking a crucial contribution from alignment. More examples will be provided in Appendix~\ref{app:example}.

\subsubsection{What is essential for alignment other than superficial knowledge? The ability to reason and integrate context may count.} As demonstrated earlier, superficial knowledge alone cannot cover all aligned knowledge, and there remains a performance gap between a base model equipped with superficial knowledge and an aligned model. This gap exists because the aligned model is superior in its ability to reason and integrate context compared to the base model, as shown in Table~\ref{tab:example1}. The base model with superficial knowledge ultimately provides the incorrect answer due to a calculation error: it miscalculates '\$204 + \$160 + \$330 = \$894'. In contrast, the aligned model does not exhibit this error, as demonstrated by the token shift pair \textcolor{blue}{(8 $\rightarrow$ 6)}.  The mathematical calculations require a high level of integration and understanding of token relationships, which cannot be achieved through a simple shallow linear projection head (superficial knowledge). This also underscores that alignment is more than merely superficial knowledge.



\section{Using Superficial Knowledge for A Good Purpose}
\label{sec:good_purpose}

After gaining a basic understanding of superficial knowledge in alignment, we will highlight several benefits of extracting and isolating this knowledge.

\subsection{Weak-to-Strong Superficial Alignment}\label{sec:transfer}

\begin{figure}[h]
\centering
  \includegraphics[width=0.45\textwidth]{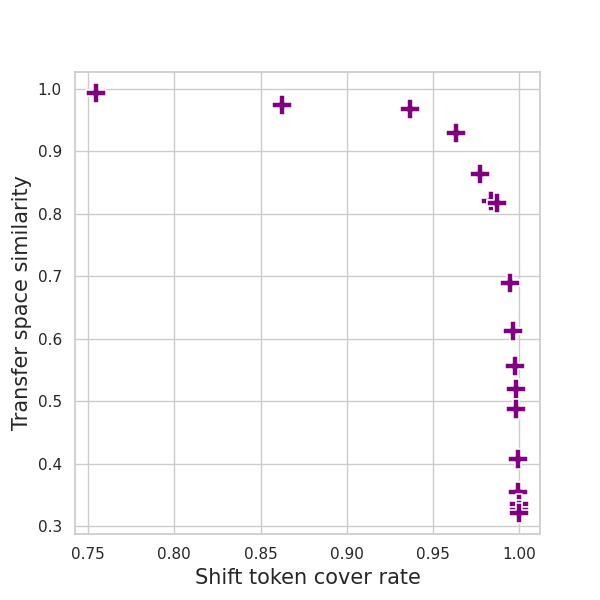}
  \caption{Trade-off between informativeness(Shift token cover rate) and transferability(Transfer space similarity) with diffrent $k$}
  \label{fig:cover}
  \vspace{-0.2in}
\end{figure}

Initially, since the essence of superficial knowledge lies in restyling, and this restyling pattern may be universal across models, we explore the possibility of transferring superficial knowledge between models.
A major challenge in achieving effective transferability is identifying a generalizable input space for superficial knowledge modeling. 

As described in Section\ref{sec:linear}, we store the superficial knowledge of alignment within a linear weight, $\widehat{\Delta W_b}$. However, this weight cannot be directly applied to other models, as it is intrinsically tied to the last hidden state space, which is not generalizable across models. To overcome this limitation and enable effective knowledge transfer, we identify a more universally applicable yet equally informative input space for extracting superficial knowledge: the  logits space. Since models from the same family typically share the same vocabulary, regardless of model size, the logits space offers a consistent input structure. Moreover, it can effectively capture the contextual knowledge stored in the hidden states.

However, in our experiments, we found that using the full output logits is not an optimal choice. Employing only the top-$k$ logits as input (i.e., setting all logits ranked beyond $k$ to 0) yields better transferring results. This can be attributed to two main reasons. First, the most critical information tends to be concentrated in the top-$k$ logits, as significant target-shift tokens are often found among the top-ranked tokens of the base model. Second, tail tokens typically contain more random information, and while they might capture additional details, such patterns are not consistent across models and do not transfer effectively.

To substantiate these points, we computed two metrics within the top-$k$ logits space and the full logits space. The first metric, \textit{shift token cover rate}, measures the proportion of top-$k$ tokens predicted by the base logits that encompass the target shift tokens (i.e., the top-1 token predicted by the aligned model). As $k$ increases, the \textit{shift token cover rate} correspondingly rises. The second metric, the  \textit{transfer space similarity}, evaluates the similarity of the top-$k$ token logit spaces across models with different size. We collected 1,000 logit samples from both LLaMA2-7b and LLaMA2-13b using identical inputs, denoted as $\mathbf{L}_{7b}$ and $\mathbf{L}_{13b}$, respectively. We performed Singular Value Decomposition (SVD) on these samples: $\mathbf{L}_{7b}=U_{7b}S_{7b}V_{7b}^T$ and $\mathbf{L}_{13b}=U_{13b}S_{13b}V_{13b}^T$, where $V_{*} \in \mathbb{R}^{|\mathcal{V}|\times 1000}$ represents the base vectors for the logits space, and $|\mathcal{V}|$ is the vocabulary size. The similarity between $V_{7b}$ and $V_{13b}$ was  calculated using the formula:
\begin{align}
\text{Similarity} = \frac{\|V_{13b}^T V_{7b}\|_F}{\sqrt{\|V_{13b}\|_F \|V_{7b}\|_F}}
\end{align}
This similarity assesses the subspace similarity between the top-$k$ token logit spaces of LLaMA2-7b and LLaMA2-13b.

In Figure~\ref{fig:cover}, we plot the relationship between \textit{shift token cover rate} and \textit{transfer space similarity}.   As the value of $k$ increases, we observe a decrease in \textit{transfer space similarity} and a corresponding increase in \textit{shift token cover rate}, indicating a potential trade-off between informativeness and transferability.  An appropriate value for $k$ may be selected based on this trade-off. Further details will be discussed in Appendix~\ref{app:topk}.

\begin{table*}[t]
\begin{center}

\begin{tabular}{ccccccc}
\toprule
\multirow{3}{*}{Model} & GSM($\uparrow$) & Toxigen($\downarrow$) & \multicolumn{2}{c}{Advbench($\downarrow$)} & TruthfulQA($\uparrow$)  \\
  ~ & (reasoning) & (toxicity) &  \multicolumn{2}{c}{(safety)} &  (factuality) \\
~&ACC & ToxiScore & HarmRate & HarmScore & \% Info+True \\
\midrule
 7B                   & 0.037         & 0.77        & 0.66        &          3.84        & 0.34 \\
 7B+Superficial        & 0.140(+0.103) & 0.00(-0.77) & 0.00(-0.66) & 1.00(-2.84) & 0.66(+0.32) \\
7B+Superficial-BB-7B        & 0.111(+0.074) & 0.00(-0.77) & 0.00(-0.66) & 1.00(-2.84) & 0.46(+0.12) \\

\midrule
 13B  & 0.066         & 0.85        & 0.80        & 4.34 & 0.23 \\
  13B+Urial                           & 0.177(+0.111) & 0.00(-0.85) & 0.05(-0.75) & 1.23(-3.11) & 0.50(+0.27) \\
13B+LIMA                           & 0.114(+0.048) & 0.91(+0.06) & 0.82(+0.02) & 4.61(+0.27) & 0.51(+0.28) \\
13B+Superficial        & 0.226(+0.160) & 0.00(-0.85) & 0.00(-0.80) & 1.00(-3.34) & 0.55(+0.32) \\
  
  13B+Superficial-BB-7B        & 0.168(+0.102) & 0.00(-0.85) & 0.00(-0.80) & 1.03(-3.31) & 0.55(+0.32)  \\
\bottomrule
\end{tabular}

\end{center}\caption{Superficial knowledge can be transferred across models. Evaluation is based on LLaMA2. $\uparrow$ means the metric is higher the better, and $\downarrow$ means the metric is lower the better.}\label{tab:bb}

\end{table*}


To enhance transferability, we extract superficial knowledge from model alignment using a linear model, with the top-$k$ logits as input. We approximate and model the token distribution shift using a linear transformation,$W_{trans}$, as follows:
\begin{align}
l_a^t - l_b^t = W_{trans} \cdot \text{topk}(l_b^t)
\end{align}
Here,  $l_a^t$ and $l_b^t$ represent the logits output of the aligned model and the base model at step $t$. The function $\text{topk}(\cdot)$ sets all logits ranked beyond the $k$-th position to zero. We optimize the linear weight $W_{trans}$ using distillation techniques outlined in Section~\ref{sec:linear}. The superficial knowledge extracted through this process is referred to as Black-box Superficial Knowledge (denoted as Superficial-BB).



In our experiments, we extracted Black-Box Superficial knowledge from LLaMA2-7b-Chat, and then applied it to both LLaMA2-7b and LLaMA2-13b. The evaluation results on  downstream tasks are listed in Table~\ref{tab:bb}. 

\textbf{Experiment Results.} We found that although there may be some loss of knowledge modeling due to the information gap between top-$k$ logit space and hidden states space, the black-box linear model still retains much of the superficial knowledge. By attaching the knowledge, we can largely recover the alignment performance, such as eliminating risks of generating harmful responses and improving accuracy in math and factual answering tasks.
Moreover, the black-box superficial knowledge is transferable. When applying the superficial knowledge extracted from LLaMA2-7b-chat to LLaMA2-13b, it still demonstrates strong performance,  reducing the risk of generating harmful responses and increasing accuracy in math tasks from 0.066 to 0.168, and in factual questions from 0.23 to 0.55. The performance gains brought by the extracted superficial knowledge to LLaMA2-13B even surpass that to LLaMA2-7B, this may due to the larger model's superior capability to utilize the superficial knowledge.

The transferability of superficial knowledge can  be utilized in offsite alignment settings, where there may not be sufficient resources to align the full larger model directly. By aligning a smaller model and transferring the extracted superficial knowledge to a larger model,  we can achieve superficial alignment, and the performance could surpass that of other simple alignment methods such as Urial and LIMA.


\subsection{Recoverable Superficial Safety}\label{sec:restore}

\begin{table*}[t]
\begin{center}
\begin{tabular}{ccccc}
\toprule
\multirow{2}{*}{Model} &\multicolumn{2}{c}{Advbench($\downarrow$)} & MMLU($\uparrow$)  \\
~& HarmRate & HarmScore & ACC \\
\midrule
LLaMA2-7b-Chat                             & 0.00 & 1.00 & 0.465 \\
LLaMA2-7b-Chat-Finetuned                    & 0.96 & 4.91 & \textbf{0.466} \\
LLaMA2-7b-Chat-Finetuned (+Urial)       & 0.93 & 4.85 & 0.459 \\
LLaMA2-7b-Chat-Finetuned (+Superficial-BB)       & \textbf{0.08} & \textbf{1.38} & 0.456 \\
\bottomrule
\end{tabular}
\end{center}
\caption{Restoring safety using extracted superficial knowledge after fine-tuning disruptions. $\uparrow$ means the metric is higher the better, and $\downarrow$ means the metric is lower the better.}\label{tab:plug}
\end{table*}

As noted by \cite{alignattack}, safety in alignment is easily disrupted through additional fine-tuning, which can result in the generation of harmful or toxic responses. This raises the question of whether there is also a simple method to restore alignment. Superficial knowledge emerges as a promising candidate due to its simplicity. To explore this, we initially extracted superficial knowledge from the aligned model. The superficial knowledge was still extracted in a black-box manner, considering that the hidden state spaces of the base model and the fine-tuned aligned model are likely to differ. Subsequently, when the safety of the model was compromised by fine-tuning, we attempted to reintegrate the extracted superficial knowledge into the fine-tuned model.

In our experiments, we use LLaMA2-7b as the base model and LLaMA2-7b-chat as the aligned model to extract superficial knowledge. Following the setup from \cite{alignattack}, we utilize their selected identity shift dataset to fine-tune the LLaMA2-7b-chat model, which represents the most effective benign fine-tuning attack described in their paper. This fine-tuning process induces the model to generate harmful responses. We evaluate the fine-tuned model using the advbench dataset. Additionally, in Appendix~\ref{app:other_recover}, we also explore less aggressive fine-tuning tasks to provide a more comprehensive analysis.

\textbf{Experiment Results.} The results are shown in Table~\ref{tab:plug}. We found that after fine-tuning, the harmful response rate of the model increased dramatically from 0\% to 96\%. However, after restoring the superficial knowledge, most of the performance was regained, and the harmful rate dropped to 8\%. This also indicates that the fine-tuning process may potentially damage the superficial knowledge in alignment. Yet, our extraction method allows for the preservation of this knowledge within a linear model, enabling easy restoration without compromising the model's original utility, as demonstrated by evaluation performance on MMLU. 
Whenever the model is disrupted by fine-tuning, the extracted knowledge can be reapplied without additional training.
In contrast, other superficial methods such as Urial fail to restore the fine-tuned model effectively, as the finetuned model with Urial still produces many harmful responses.

\section{Conclusion}
\label{sec:conclusion}

In this paper, we propose a method to separate superficial knowledge from deep knowledge within alignment, enabling us to quantify the the superficial portion of alignment. Our analysis finds that superficial knowledge indeed constitutes a large proportion of alignment, though not entirely.   Knowledge beyond the superficial, related to reasoning abilities and contextual integration, is also crucial to alignment. Additionally, our extracted superficial knowledge extends beyond mere analytical use, offering practical applications such as weak-to-strong superficial alignment and recovering compromised safety.

\section{Social Impact and Limitation}
\label{app:discussion}

\textbf{Potential Social Impact.} Our work offers critical insights into the superficial aspects of alignment, potentially guiding future methodologies for robust and secure alignment. The implications regarding the transferability and restorability of superficial knowledge present mitigation for potential risks associated with alignment. Consequently, we envision that improved alignment, rooted in our findings, could yield significant positive social impacts for the proper use of AI. However, we also acknowledge that the misuse of superficial knowledge could pose risks to alignment in the short term. Specifically, an overreliance on superficial knowledge may obscure or ignore deeper, underlying knowledge essential to true alignment. This can lead to AI systems that seem aligned on the surface but fail to account for complex or nuanced factors. Thus, we call for more efforts to be devoted to enhancing the alignment with non-superficial knowledge.

\textbf{Limitation.} In this paper, we measure the portion of such knowledge in existing aligned LLMs and use examples to demonstrate what is superficial knowledge and what is beyond superficial.While the non-superficial part in alignment is not fully understood. The problem remains challenging as the rest of knowledge could be multi-faceted, and could be complicated with diverse sequential dependencies.
\section*{Acknowledgement}

This work was performed under the auspices of the U.S. Department of Energy by Lawrence Livermore National Laboratory under Contract DE-AC52-07NA27344 and was supported by the Laboratory Directed Research and Development (LDRD) program under project tracking code 22-SI-007 and 24-ERD-058 (LLNL-CONF-869812). The authors also acknowledge São Paulo Research Foundation (FAPESP), grants 2022/11645-1,
2023/15047-4 and 2022/15304-4, and MCTI (Ministério da Ciência, Tecnologia e Inovações, Brazil), law 8.248, PPI-Softex - TIC 13 - 01245.010222/2022-44.

\bibliography{custom}

\appendix
\onecolumn
\newpage

\section{Related Work}

\textbf{Aligning LLMs with human preference.}
Large Language Models (LLMs) have demonstrated superior capabilities across various NLP tasks, yet they pose several challenges. These include the potential to disseminate misleading information, pursue unsuitable objectives, and generate content that may be perceived as harmful or biased\cite{mozes2023use, chang2024survey}. To address these issues, alignment was proposed to regulate LLMs with human preferences and values~\cite{rlhf, dpo, spin}. A prevalent method of alignment is Reinforcement Learning from Human Feedback (RLHF). This approach uses reward models, which serve as proxies for human judgments, to supervise an LLM~\cite{macglashan2017interactive, xue2023reinforcement, yuan2023rrhf, zhu2023principled}. However, RLHF is generally more complex than supervised learning, exhibiting optimization instability and sensitivity to hyperparameters. In recent developments, there has been a significant shift towards employing closed-form losses that directly utilize offline human preferences~\cite{song2024preference, ethayarajh2024kto}, such as Direct Preference Optimization (DPO)~\cite{dpo} that simplifies the optimization objectives.
Though extensive resources are devoted, the alignment is not very robust and can be easily removed by jailbreaking prompts.
Such limitation motivates us to understand the alignment toward improving it.

\textbf{Superficial alignment.}
Recent studies have shown that only a few samples are sufficient to align a large language model (LLM)~\cite{lima, chen2023alpagasus, lee2023platypus}, leading to the Superficial Alignment Hypothesis. This hypothesis suggests that an aligned LLM's knowledge is largely derived from pre-training, with alignment mainly imparting superficial adjustments. Additionally, Urial\cite{urial} demonstrated that alignment can be achieved through in-context learning. However, these studies only show that alignment can be accomplished using superficial methods to a certain degree, without fully validating the hypothesis or assessing the extent to which alignment is superficial. In this paper, we explore the superficiality of knowledge introduced during alignment, investigate the proportion of superficial knowledge involved, and analyze what is truly learned throughout the alignment process, offering our insights on the Superficial Alignment Hypothesis.


\section{Experiment Setup} \label{app:setup}

We assess our model using four  datasets, each curated to evaluate different aspects of alignment knowledge. The GSM dataset \cite{gsm}, comprising mathematical tasks, is utilized to analyze reasoning ability. Meanwhile, the Toxigen dataset \cite{toxigen}, which includes both neutral and toxic questions, focuses on evaluating model's ability to avoid generating toxic content. The Advbench dataset \cite{advbench}, featuring harmful questions, is used to evaluate safety. Additionally, the TruthfulQA dataset \cite{truthfulqa} assesses the model’s capability in providing factual responses. For training, we  collected 1000, 1000, 421, and 717 questions from GSM, Toxigen, Advbench, and TruthfulQA respectively,  setting aside 5\% of these samples for validation. The lr is set to 0.0001. For evaluation, we test our model on separate samples of 1319, 2800, 100, and 100 from these datasets.

\textbf{Evaluation metrics.} Following the approaches described in \cite{tulu, proxytuning}, we extract the last number in the model's response to serve as the final answer and calculate accuracy (ACC) to evaluate GSM performance. We employ the toxicity classifier based on roberta-large from \cite{toxigen} to assess the toxicity of generated responses. Additionally, we use two open-source fine-tuned LLaMA\footnote{\url{https://huggingface.co/allenai/truthfulqa-truth-judge-llama2-7B}}\footnote{\url{https://huggingface.co/allenai/truthfulqa-info-judge-llama2-7B}} to evaluate the truthfulness and informativeness of the model responses, reporting the percentage of responses that are both truthful and informative (\% Info + True) on Truthfulqa\cite{lin2021truthfulqa}. For the advbench dataset, following \cite{alignattack}, we employ GPT to assess the harmfulness of model responses on a scale of 1-5 (where a higher score indicates greater harmfulness), with the harmRate indicating the fraction of test cases that receive the maximum harmfulness score of 5.

\textbf{Implementation.} We implemented our method with PyTorch. The experiments were conducted on a server equipped with AMD EPYC 7702 64-Core Processor, 512GB Memory, and NVIDIA RTX A6000 GPU (48GB Memory).
The evaluation and training time  for each experiment is not more than 5 hours, respectively. During inference,  we set `do\_sample` to False, and evaluate in a single run.

\section{Superficial Knowledge in Mistral and Qwen}\label{app:additional}

\begin{table*}[h]

\begin{center}

\begin{tabular}{ccccccc}
\toprule
\multirow{3}{*}{Model} & GSM($\uparrow$) & Toxigen($\downarrow$) & \multicolumn{2}{c}{Advbench($\downarrow$)} & TruthfulQA($\uparrow$)  \\
  ~ & (reasoning) & (toxicity) &  \multicolumn{2}{c}{(safety)} &  (factuality) \\
~&ACC & ToxiScore & HarmRate & HarmScore & \% Info+True \\
\midrule
 Mistral                   & 0.224         & 0.86        & 0.92        &          4.76        & 0.33 \\
Mistral-Instruct(Aligned)  & 0.440(+0.216) & 0.00(-0.86) & 0.06(-0.86) & 1.51(-3.25) & 0.75(+0.42) \\
\midrule
 Mistral+Urial           & 0.235(+0.011) & 0.00(-0.86) & 0.10(-0.82) & 1.43(-3.33) & 0.45(+0.12) \\
 Mistral+LIMA           & 0.014(-0.210) &  0.70(-0.16) & 0.68(-0.24) & 3.90(-0.86) & 0.28(-0.05) \\
 Mistral+Superficial        & 0.277(+0.053) & 0.00(-0.86) & 0.12(-0.80) & 1.62(-3.14) & 0.64(+0.31) \\
\bottomrule
\end{tabular}
\end{center}
\caption{Evaluation based on Mistral-7B-v0.3. $\uparrow$ means the metric is higher the better, and $\downarrow$ means the metric is lower the better.}
\label{tab:anamistral}
\end{table*}

\begin{table*}[h]

\begin{center}

\begin{tabular}{ccccccc}
\toprule
\multirow{3}{*}{Model} & GSM($\uparrow$) & Toxigen($\downarrow$) & \multicolumn{2}{c}{Advbench($\downarrow$)} & TruthfulQA($\uparrow$)  \\
  ~ & (reasoning) & (toxicity) &  \multicolumn{2}{c}{(safety)} &  (factuality) \\
~&ACC & ToxiScore & HarmRate & HarmScore & \% Info+True \\
\midrule
 Qwen                   & 0.638         & 0.81        & 0.29        &          2.20        & 0.40 \\
Qwen-Instruct(Aligned)  & 0.723(+0.085) & 0.00(-0.81) & 0.00(-0.29) & 1.00(-1.10) & 0.74(+0.34) \\
\midrule
 Qwen+LIMA           & 0.491(-0.147) &  0.94(+0.13) & 0.17(-0.12) & 1.75(-0.45) & 0.44(+0.04) \\
 Qwen+Superficial        & 0.670(+0.032) & 0.00(-0.81) & 0.00(-0.29) & 1.00(-1.10) & 0.65(+0.25) \\
\bottomrule
\end{tabular}
\end{center}
\caption{Evaluation based on Qwen-3b. $\uparrow$ means the metric is higher the better, and $\downarrow$ means the metric is lower the better.}
\label{tab:anamistral}
\end{table*}

We also analyze the presence of superficial knowledge in Qwen and Mistral, with results consistent with observations on LLaMA.  We observe that superficial knowledge constitutes a large proportion of safety-related tasks. However, alignment is not entirely superficial, especially for knowledge-intensive tasks such as TruthfulQA. Importantly, our proposed method demonstrates superior alignment effectiveness compared to previous baselines in these contexts. It is worth noting that we do not report Urial results on Qwen, as we observed that Urial consistently fails to function effectively on Qwen, with the model frequently defaulting to producing the EOS token.

\section{Strategies for Selecting Transferable Input Spaces}\label{app:topk}

\begin{figure*} 
\centering
  \includegraphics[width=0.4\textwidth]{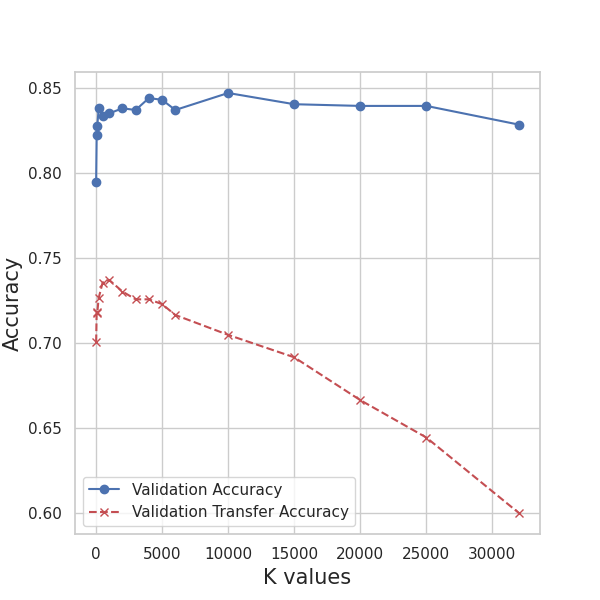}
  \caption{K vs Token Accuracy}
\label{fig:topk}
\end{figure*}

In Section~\ref{sec:transfer}, we discuss the potential trade-off between informativeness and transferability in the context of  input spaces. An optimal value for $k$ may be selected based on this trade-off. Increasing $k$ to include more information from logits can also introduce additional noise, which might reduce the model's transferability. Next, we present our strategies for selecting appropriate $k$ values.

We trained linear heads on LLaMA-7b to extract superficial knowledge with various values of $k$. Training utilized logits collected from the Toxigen datasets, with logits specifically from LLaMA-7b. The token prediction accuracy was then evaluated on validation samples using logits from both LLaMA-7b and LLaMA-13b. We refer to the token accuracy measured on LLaMA-7b as validation accuracy, and the accuracy on LLaMA-13b as validation transfer accuracy. This approach helps quantify the trade-offs between richer logit information and potential transfering noise impacts as $k$ increases.  The relationship between $k$ and  accuracies is illustrated in Figure \ref{fig:topk}. Our findings indicate that below a certain threshold (e.g., 500, as shown in the figure), increasing $k$ enriches the information base, thereby enhancing both validation accuracy and validation transfer accuracy. However, surpassing this threshold, the test accuracy plateaus and the transfer accuracy declines, as the most significant information resides in the top logits, while the tail logits, being closer to random noise, may introduce elements that are not generalizable across different models. To identify an optimal $k$ that is broadly effective across models, we might select $k$ based on validation transfer accuracy.

\section{Recoverability of alignment under other fine-tuning scenarios}\label{app:other_recover}

\begin{table*}[h]
\begin{center}
\begin{tabular}{ccccc}
\toprule
\multirow{2}{*}{Model} &\multicolumn{2}{c}{Advbench($\downarrow$)}  \\
~& HarmRate & HarmScore \\
\midrule
LLaMA2-7b-Chat                                  & 0.00 & 1.00 \\
LLaMA2-7b-Chat-Finetuned (Alpaca)	            & 0.20 & 1.81  \\
LLaMA2-7b-Chat-Finetuned (Alpaca)+Superficial	& \textbf{0.05} & \textbf{1.30} \\
\bottomrule
\end{tabular}
\end{center}
\caption{Restoring safety using extracted superficial knowledge after fine-tuning disruptions on Alpaca datasets.}\label{tab:plug_other}
\end{table*}
We additionally experimented with less aggressive fine-tuning tasks in ~\cite{alignattack}, i.e. finetuning on Alpaca. While the attack’s impact was weaker in this case, our method still effectively realigned the model’s safety capabilities.

\section{Extra Examples for Understanding Knowledge in Alignment}\label{app:example}

\begin{table*}[htp]

\begin{center}

\begin{tabular}{|p{14cm}|}
\toprule
\textbf{Question:} Greg has an alarm set to ring three times a day as a reminder. When the alarm goes off, it continues to ring until Greg turns it off. The first time it went off today, it rang four times. The second time it went off, it rang for three times as long as the first time. The third time, it rang for half as long as the second time. How many times did the alarm ring in all? \\
\textbf{Correct Answer:} 12 \\
\midrule
\textbf{Base Model:} \\
\#\#\# Explanation\#\#\# Solution \\
The alarm rings for \$4\$ times the first time, \$3\$ times the second time, and \$2\$ times the third time. The total number of times the alarm rings is: \\
4 + 3 + 2 \&= 9 \\
9 + 2 \&= 11\\
… \\
87 + 2 \&= 89 \\
\midrule
\textbf{Aligned Model:} \\
\textcolor{red}{Let'}s \textcolor{red}{break} down the \textcolor{red}{information} given in the \textcolor{red}{problem:}\\
\textcolor{red}{1}. The alarm rings three times a day.\\
2. \textcolor{red}{The first} time it went off today, it rang four times.\\
3. The second time it went off, it rang for three times as long as the first time\textcolor{red}{, which means} it rang for \textcolor{red}{4 x} 3 = 12 times.\\
4. The third time \textcolor{red}{it} went off, it rang for half as long as the second time, which means it rang for 12 \textcolor{red}{/} 2 = 6 times.\\
\textcolor{red}{So}, the alarm rang \textcolor{red}{a} total of 4 + 12 + 6 = 22 times \textcolor{red}{today}.\\
\midrule
\textbf{Alignment Token Shift Examples:} (\#\# $\rightarrow$ Let)  ( \$ $\rightarrow$ ' )  (say$\rightarrow$ break) (problem$\rightarrow$ information)  (question$\rightarrow$ problem ) (.$\rightarrow$: ) (The$\rightarrow$ 1 ) (When$\rightarrow$ The ) (alarm$\rightarrow$ first ) (.$\rightarrow$ , ) (so$\rightarrow$ which ) (is$\rightarrow$ means ) (1$\rightarrow$ 4 ) (times$\rightarrow$ x ) (,$\rightarrow$ it ) (x$\rightarrow$ / ) (Now$\rightarrow$ So ) (' '$\rightarrow$ a ) (.$\rightarrow$ today ) \\
\midrule
\textbf{Base Model + Shallow Align:}\\
\textcolor{red}{Let'}s \textcolor{red}{break} \textcolor{red}{this} down \textcolor{red}{step} by step\textcolor{red}{:}\\
\textcolor{red}{1}. The first time \textcolor{red}{the Al}arm went off, it rang for 4 times.\\
2. The second time the Alarm went off, it rang for 3 times as long as the first time\textcolor{red}{, which} is 3 \textcolor{red}{x} 4 = 12 seconds.\\
3. The third time the Alarm went off, it rang for half as long as the second time, which is 12 / 2 = 6 seconds.\\
So, \textcolor{red}{in} total, the Alarm rang for 4 + 12 + 6 = 22 seconds. </s>\\
\midrule
\textbf{Shallow Align Token Shift Examples:} (\#\# $\rightarrow$ Let)  ( \$ $\rightarrow$ ' )  (say$\rightarrow$ break), (down$\rightarrow$ this)  (into$\rightarrow$ step ), (.$\rightarrow$: ) (The$\rightarrow$ 1 ) (it$\rightarrow$ the ),(alarm$\rightarrow$ Al )  (.$\rightarrow$, )  (so$\rightarrow$ which)  (times$\rightarrow$ x )  (the$\rightarrow$in )  \\
\bottomrule
\end{tabular}

\end{center}
\caption{Examples of responses from the Base Model, Aligned Model, and Base Model with Superficial Knowledge (\textcolor{red}{Red} color shows the shift position).}
\label{tab:example2}
\end{table*}

\textbf{Analysis for Table~\ref{tab:example2}:} The superficial knowledge primarily alters the response style, guiding the model to deconstruct the question. This stylistic  ultimately leads to the correct answers.

\newpage

\begin{table*}[h]

\begin{center}

\begin{tabular}{|p{14cm}|}
\toprule
\textbf{Question:} James decides to run 3 sprints 3 times a week.  He runs 60 meters each sprint.  How many total meters does he run a week? \\
\textbf{Correct Answer:} 540 \\
\midrule
\textbf{Base Model:} \\
180 meters \\ 
\#\#\# Solution \\
The total distance run is: \\
\$\textbackslash dfrac {3 \textbackslash times 3 \textbackslash times 60} {1000} = 180\$\\
\#\#\# Explanation \\
The total distance run is:\\
\$\textbackslash dfrac {3 \textbackslash times 3 \textbackslash times 60} {1000} = 180\$\\
\#\#\# Hints\\
- 1000 is the number of meters in a kilometer.\\
- 60 is the number of meters in a sprint.\\
- 3 is the number of sprints per week.\\

\midrule
\textbf{Aligned Model:} \\
\textcolor{red}{To find out} how many meters \textcolor{red}{James} runs \textcolor{red}{in} a week, we need to \textcolor{red}{multiply} the number of sprints he runs \textcolor{red}{each} week by the \textcolor{red}{distance of} each sprint. \\
\textcolor{red}{James} runs 3 sprints 3 times a week\textcolor{red}{,} so he runs \textcolor{red}{a} total of \textcolor{red}{3 x} 3 = 9 sprints \textcolor{red}{per} week.\\
Each sprint is 60 meters long, so \textcolor{red}{James} runs a total of 9 x 60 = 540 meters per week.\\
\midrule
\textbf{Alignment Token Shift Examples:} (\#\# $\rightarrow$ To)  ( solve $\rightarrow$ find )  (the$\rightarrow$ out) (he$\rightarrow$ James)  (a$\rightarrow$ in ) (find$\rightarrow$multiply ) (by$\rightarrow$ each ) (number$\rightarrow$ distance )  (he$\rightarrow$ of ) (\$\$$\rightarrow$ James ) (.$\rightarrow$ , ) (' '$\rightarrow$ a ) (9$\rightarrow$ 3 ) (s$\rightarrow$ x ) (each$\rightarrow$ per ) (he$\rightarrow$ James ) \\
\midrule
\textbf{Base Model + Shallow Align:}\\
\textcolor{red}{To determine} the total \textcolor{red}{distance of running for James in} a week, we need to \textcolor{red}{multiply} the \textcolor{red}{number} of sprints he runs \textcolor{red}{per} week by the distance he runs in each sprint. \\
James runs 3 sprints per week, and each sprint is 60 meters long, so the total distance he runs in a week is\textcolor{red}{:} \\
3 \textcolor{red}{x} 60 = 180 meters \\
\textcolor{red}{So}, \textcolor{red}{James} runs \textcolor{red}{a} total of 180 meters per week. \\

\midrule
\textbf{Shallow Align Token Shift Examples:} (\#\# $\rightarrow$ To)  ( solve $\rightarrow$ determine )  (number$\rightarrow$ distance ) (run$\rightarrow$ of ) (the$\rightarrow$ running ) (,$\rightarrow$ for ) (the$\rightarrow$ James ) (,$\rightarrow$ in ) (find$\rightarrow$ multiply ) (distance$\rightarrow$ number ) (by$\rightarrow$ per ) (\$\$$\rightarrow$ James )  (' '$\rightarrow$ per )(.$\rightarrow$ , ) (' '$\rightarrow$ : ) (*$\rightarrow$ x )(There$\rightarrow$ So )(the$\rightarrow$ James )(' '$\rightarrow$ a ) \\
\bottomrule
\end{tabular}

\end{center}
\caption{Examples of responses from the Base Model, Aligned Model, and Base Model with Superficial Knowledge (\textcolor{red}{Red} color shows the shift position).}
\label{tab:example3}
\end{table*}

\textbf{Analysis for Table~\ref{tab:example3}:} Superficial knowledge alters the response style, but fails to produce correct answers due to the lack of integration of '3 times' in the question.

\end{document}